# Path tracking and stabilization for a reversing general 2-trailer configuration using a cascaded control approach

Niclas Evestedt[1], Oskar Ljungqvist[1], Daniel Axehill[1]

***Abstract*— In this paper a cascaded approach for stabilization and path tracking of a general 2-trailer vehicle configuration with an off-axle hitching is presented. A low level Linear Quadratic controller is used for stabilization of the internal angles while a pure pursuit path tracking controller is used on a higher level to handle the path tracking. Piecewise linearity is the only requirement on the control reference which makes the design of reference paths very general. A Graphical User Interface is designed to make it easy for a user to design control references for complex manoeuvres given some representation of the surroundings. The approach is demonstrated with challenging path following scenarios both in simulation and on a small scale test platform.**

## I. INTRODUCTION

Many Advanced Driver Assistance Systems (ADAS) have been introduced during the last decade. Historically the focus has been to increase safety with systems like Lane Keep Assist (LKA), Adaptive Cruise Control (ACC) and Automatic Breaking, however in recent years systems that help the driver to perform complex tasks, such as parallel parking and reversing with a trailer, have been introduced [1]. Reversing with a trailer is known to be a task that needs a fair amount of skill and training to perfect and an inexperienced driver will have problems already performing simple tasks such as reversing in a straight line or make a simple turn around an obstacle. To relieve the driver in such situations, trailer assist systems have been developed that stabilize the trailer around a reference that the driver can specify from a control knob. The trailer assist systems have been released to the passenger car market but an even greater challenge arises when reversing a truck with a dolly steered trailer. This introduces another degree of freedom making it virtually impossible for a driver, without extensive training, to control. In this paper we present a cascaded control scheme using an LQ-controller, based on the work in [2], to stabilize the vehicle configuration around an equilibrium point and then use a pure-pursuit path tracking controller to make the vehicle configuration follow a piecewise linear reference path. We also present an intuitive interface where a user can specify a path from start to goal by hand and then let the system execute that path automatically.

### A. Related work

The nonlinear dynamics of a standard trailer configuration with the hitch connection in the center of the rear axle are well understood and the derivation of the equations for an n-trailer configuration can be found in [3]. In [4], [5], [6] the flatness property of this system is used and controllers using feedback linearization are designed, [6] and [5] also demonstrate the feasibility of the controllers using some 1-trailer lab experiments. However, the assumption that the hitch connection is longitudinally centered at the rear axle center does not hold for passenger cars nor for the truck that will be used in this work. The nonlinear dynamics of the general n-trailer system, where no assumptions are made on the position of the hitching point are derived in [7]. Input-output linearization is used in [1] to derive a controller for the 1-trailer system with off-axle hitching that stabilize the trailer's driving curvature, they also derive a path tracking controller and test the controllers with good results on a real car test platform. Although these results are very encouraging it is shown in [8] and [9] that trailer configurations with more than one trailer are not input-output linearizable. To overcome this [10] introduce what they call a *"ghost vehicle"* that should be exact linearizable and have similar behaviour to the original model. A controller can then be designed using the exact linearization techniques for the ghost model and then apply the controller to the original system. An LQ based approach for reversing the general 2-trailer system with off-axle hitching is presented in [2]. The system is linearized around an equilibrium point and the linearized system is used to design an LQ control scheme. The work focuses on the stabilization of the internal angles but does not look into path tracking.

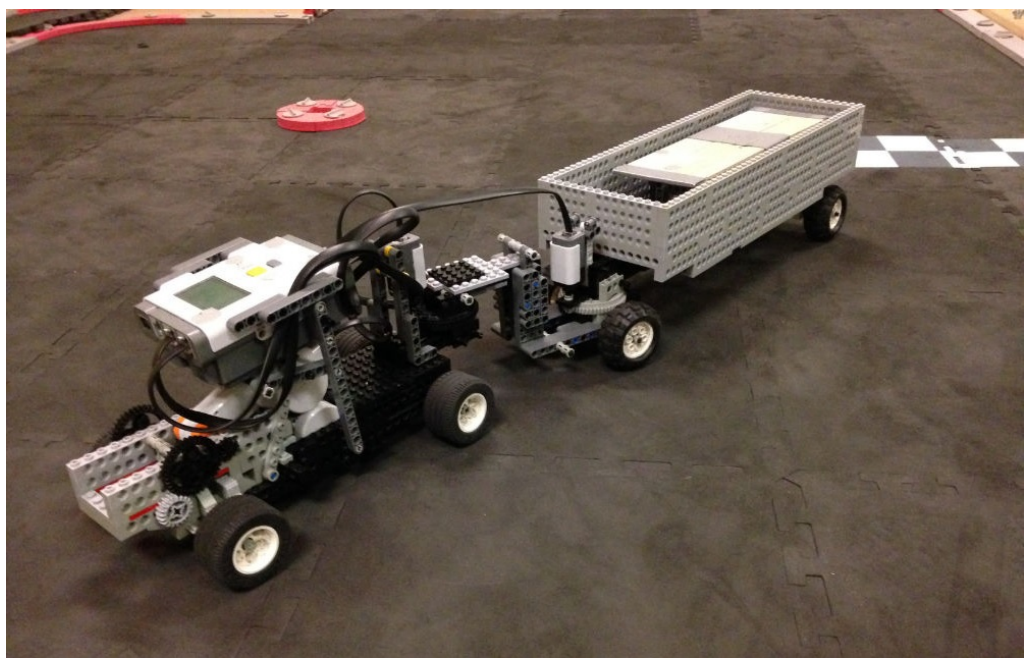

Fig. 1: Truck and trailer system used as a test platform for evaluation experiments. The truck has been built with LEGO NXT and fitted with angle sensor for dolly and hitch angles.

This work presents results from the thesis [11], where results from [2] are used to stabilize the internal angles of the system and then a pure pursuit path tracking controller is designed for the stabilized system in a similar way as [12]

*The research leading to these results has been carried out within the iQMatic project funded by FFI/VINNOVA.

[1]Division of Automatic Control, Linköping University, Sweden, (e-mail: {niclas.evestedt, oskar.ljungqvist, daniel.axehill}@liu.se)

and [13]. Finally we present an intuitive user interface that can be used to perform complex manoeuvres from a user-defined start position to a user-defined goal position. The system is then validated both in simulation and on a small-scale test platform.

The outline of the remainder of the paper is as follows: In Section II the nonlinear equations that are used to model the system are presented. In Section III the LQ controller used will be explained and in Section IV the pure pursuit path tracking controller will be discussed. Section V presents the user interface and Section VI presents the small scale test platform that was used for experiments. Finally, Section VII and Section VIII present the results and conclusions, respectively.

## II. System dynamics

In this section we present the model we have used to describe the general 2-trailer system with an off-axle hitching on the pulling vehicle. A schematic overview of the configuration is shown in Fig 2. The generalized coordinates used to model the system are, $\mathbf{p} = [x_3, y_3, \theta_3, \beta_3, \beta_2]^T$ where $x_3, y_3$ are the position of the rear axle center of the trailer, $\theta_3$ is the heading of the trailer, $\beta_3$ is the relative angle between the trailer and the dolly and $\beta_2$ is the relative angle between the dolly and the truck. The parameters $L_3, L_2, L_1$ are the distances between the axle center for the trailer to the axle center for the dolly, the axle center for the dolly to the off-axle hitch connection for the truck and the distance between the axle centers for the truck, respectively. $M_1$ is the off-axle hitch length for the truck and $\alpha$ is the steering angle. The dynamic model for a general n-trailer system was derived in [7] and the following equations for the special 2-trailer case was presented in [2]:

$$\dot{x}_3 = v\cos\beta_3\cos\beta_2\left(1+\frac{M_1}{L_1}\tan\beta_2\tan\alpha\right)\cos\theta_3 \quad (1)$$

$$\dot{y}_3 = v\cos\beta_3\cos\beta_2\left(1+\frac{M_1}{L_1}\tan\beta_2\tan\alpha\right)\sin\theta_3 \quad (2)$$

$$\dot{\theta}_3 = v\frac{\sin\beta_3\cos\beta_2}{L_3}\left(1+\frac{M_1}{L_1}\tan\beta_2\tan\alpha\right) \quad (3)$$

$$\dot{\beta}_3 = v\cos\beta_2\left(\frac{1}{L_2}\left(\tan\beta_2-\frac{M_1}{L_1}\tan\alpha\right)-\frac{\sin\beta_3}{L_3}\left(1+\frac{M_1}{L_1}\tan\beta_2\tan\alpha\right)\right) \quad (4)$$

$$\dot{\beta}_2 = v\left(\frac{\tan\alpha}{L_1}-\frac{\sin\beta_2}{L_2}+\frac{M_1}{L_1L_2}\cos\beta_2\tan\alpha\right) \quad (5)$$

where $v$ is the longitudinal velocity at the rear axle for the truck. The model is valid under the no slip condition. Since our application only concerns maneuvers at lower speeds this is a feasible assumption.

### A. Linearization

To fit into the LQ-framework used in the next section a linearized system model needs to be derived. When driving forward, the system is stable but in reverse $(v < 0)$ the system become unstable. Since v enters linearly in (1)-(5) it only affect the system as a time scaling [11] and will, theoretically, have no effect on the controller design. However, in practice backlash and unmodeled dynamics in the steering mechanism will limit the maximum speed for stability. Since a cascaded approach is used in our approach the slower states, $x_3, y_3$ and $\theta_3$ will be controlled by a high level controller. Hence, only (4) and (5) for the states $\beta_2$ and $\beta_3$ are considered in the linearization and stabilization of the model for the low level controller.

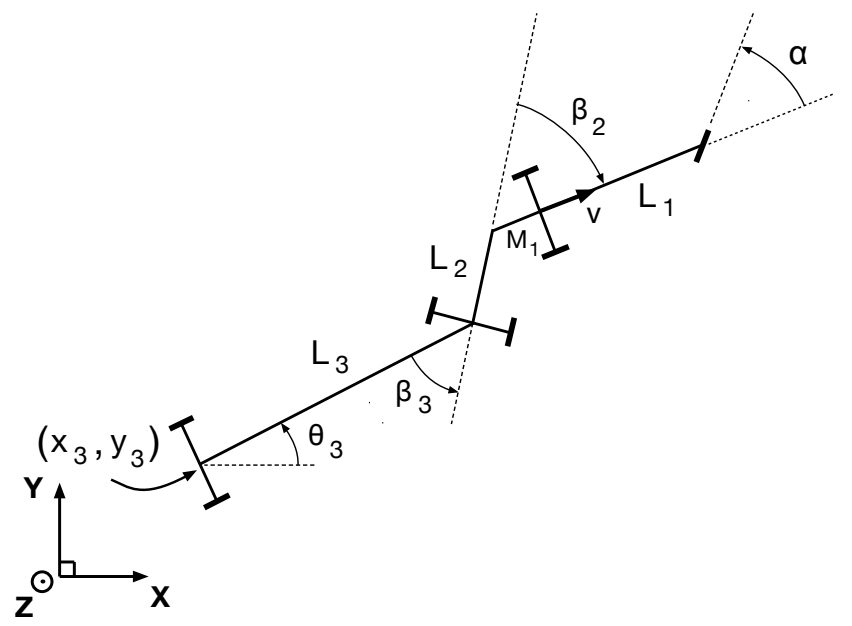

Fig. 2: Schematic view of the configuration used to model the two-trailer system.

Given a constant steering angle, $\alpha_e < \alpha_{max}$, where $\alpha_{max}$ is defined in (10), there exists a stationary equilibrium configuration where $\dot{\beta}_2$ and $\dot{\beta}_3$ equal zero. At this equilibrium point the configuration will travel along a circle with a radius determined by $\alpha_e$ as depicted in Fig. 3. From Fig. 3 the angles, $\beta_{2e}$ and $\beta_{3e}$, at this equilibrium can be determined by basic trigonometry and gives the following relations:

$$\beta_{3e} = \operatorname{sign}(\alpha_e)\arctan\left(\frac{L_3}{R_3}\right) \quad (6)$$

$$\beta_{2e} = \operatorname{sign}(\alpha_e)\left(\arctan\left(\frac{M_1}{R_1}\right)+\arctan\left(\frac{L_2}{R_2}\right)\right) \quad (7)$$

where $R_1 = L_1/|\tan\alpha_e|$, $R_2 = \sqrt{R_1^2+M_1^2-L_2^2}$, $R_3 = \sqrt{R_2^2-L_3^2}$.

With $f(\boldsymbol{\beta}) = [\dot{\beta}_3, \dot{\beta}_2]^T$ and linearizing around this reference point and letting $\boldsymbol{\beta} = [\beta_3, \beta_2]^T$ we get

$$\dot{\boldsymbol{\beta}} = \bar{A}(\boldsymbol{\beta}-\boldsymbol{\beta_e})+\bar{B}(\alpha-\alpha_e) \quad (8)$$

where

$$\bar{A} = \left.\frac{\partial f}{\partial \boldsymbol{\beta}}\right|_{\boldsymbol{\beta_e},\alpha_e} \text{ and } \bar{B} = \left.\frac{\partial f}{\partial \alpha}\right|_{\boldsymbol{\beta_e},\alpha_e} \quad (9)$$

The set of equilibrium configurations reaches its limit when the traveling circle for the rear axle of the trailer collapses to a point at the center of the rear axle. This happens when $\beta_{3e} = \pi/2$ which directly gives $R_3 = 0$ and $R_2 = L_3$ at this point. Inserting this in the relation for $R_2$ and solving for $\alpha_e$ we get the maximum steering angle where the linearization is valid for a given parameter set $L_1$, $L_2$, $L_3$ and $M_1$.

$$\alpha_{max} = \arctan\left(\sqrt{\frac{L_1^2}{L_3^2+L_2^2-M_1^2}}\right) \quad (10)$$

This linearized model can now be used for control design and this is further explained in the next section.

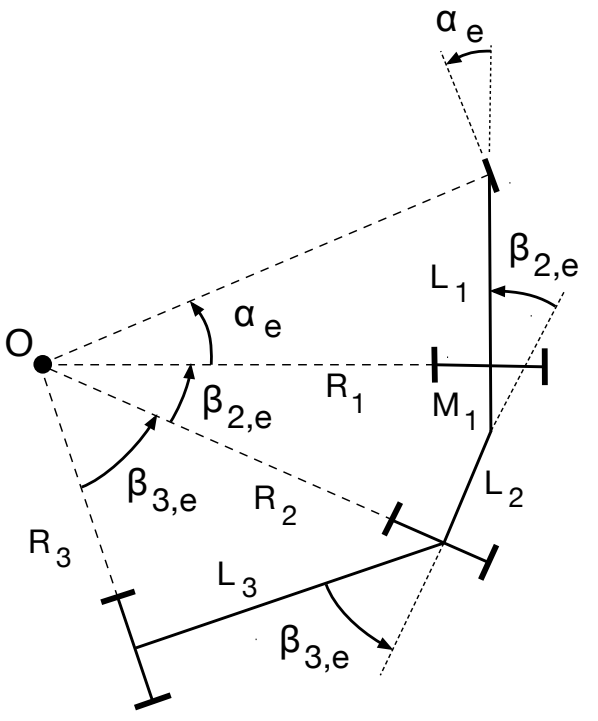

Fig. 3: Stationary equilibrium point for steering angle $\alpha_e$. The system will travel in a circular path with a radius determined by the geometry and $\alpha_e$.

## III. Stabilization

For the low level stabilization of $\beta_2$ and $\beta_3$ a state feedback controller, using LQ-techniques similar to what was done in [2] but extended with gain scheduling, is designed. The controller is designed around an equilibrium point using (8) for the linearized model. To account for model variation due to the choice of equilibrium point, $\alpha_e$, a gain scheduled approach depending on $\alpha_e$ is used. Using $\alpha_e$ as a reference results in a controller with the following structure

$$\alpha = \alpha_e - L(\alpha_e)\left(\boldsymbol{\beta} - \boldsymbol{\beta}_e(\alpha_e)\right) \tag{11}$$

where $\alpha_e$ is the desired linearization point, $L(\alpha_e)$ is the controller gain at this point, $\boldsymbol{\beta}$ the current measured angles and $\boldsymbol{\beta}_e$ is the steady state values for $\beta_2$ and $\beta_3$ given by (6) and (7) for a given $\alpha_e$.

LQ design is a well known method and thoroughly documented in control literature and we refer to [14] for details. Given a linearization point $\alpha_e$ and the linearized model in (8), the LQ design method finds the optimal gain, $L(\alpha_e)$ minimizing the cost function

$$\mathcal{J} = \int_0^\infty \left(\bar{\boldsymbol{\beta}}^T Q \bar{\boldsymbol{\beta}} + \bar{\alpha}^2\right) dt \tag{12}$$

where $\bar{\boldsymbol{\beta}} = \boldsymbol{\beta} - \boldsymbol{\beta}_e$, $\bar{\alpha} = \alpha - \alpha_e$ and $Q$ is a design parameter. By solving the problem for different linearization points, $\alpha_e$, in the range given by the linearization limits given by (10), $L(\alpha_e)$ can be obtained as show in Fig. 4. For the high level path follower, that will be further explained in the next section, we want to be able to control $\beta_3$ instead of $\alpha$. To achieve this we introduce $\beta_{3e}$ as the reference to the controller by deriving a pre-compensation link from $\beta_{3e}$ to $\alpha_e$. From (6) and the definitions for $R_1$, $R_2$ and $R_3$ we get

$$\alpha_e = \arctan\left(\frac{L_1 \operatorname{sign}(\beta_{3e})}{\sqrt{L_3^2\left(1+\frac{1}{\tan^2\beta_{3e}}\right)+L_2^2-M_1^2}}\right) \tag{13}$$

## IV. Path tracking

In this section the high level path tracking controller is introduced. When the internal angles $\beta_2$ and $\beta_3$ are stabilized by the LQ controller a pure pursuit path follower is used as a high level controller to stabilize the system around a reference path. The pure pursuit controller has successfully been used for path tracking for mobile platforms before, e.g [15]. In forward motion the controller has direct control of the steering angle $\alpha$ and the anchor point of the look-ahead circle is set at the rear axle center of the pulling vehicle. In backwards motion however the look-ahead circle anchor point, $P^*$, is set in the center of the rear axle of the last trailer and the pure pursuit controller gives the reference $\beta_{3d}$ to the low level stabilizing controller as depicted in Fig. 5. A piecewise linear reference path is given as input and by calculating the intersection of the look-ahead circle with radius, $L_r$, and the line segment between two points on the reference path, the error heading, $\theta_e$, can be calculated. The control law for $\beta_{3d}$ that will drive the vehicle along a circle to the look-ahead point can now be found using basic geometry giving

$$\beta_{3d} = -\arctan\left(\frac{2L_3\sin\theta_e}{L_r}\right) \tag{14}$$

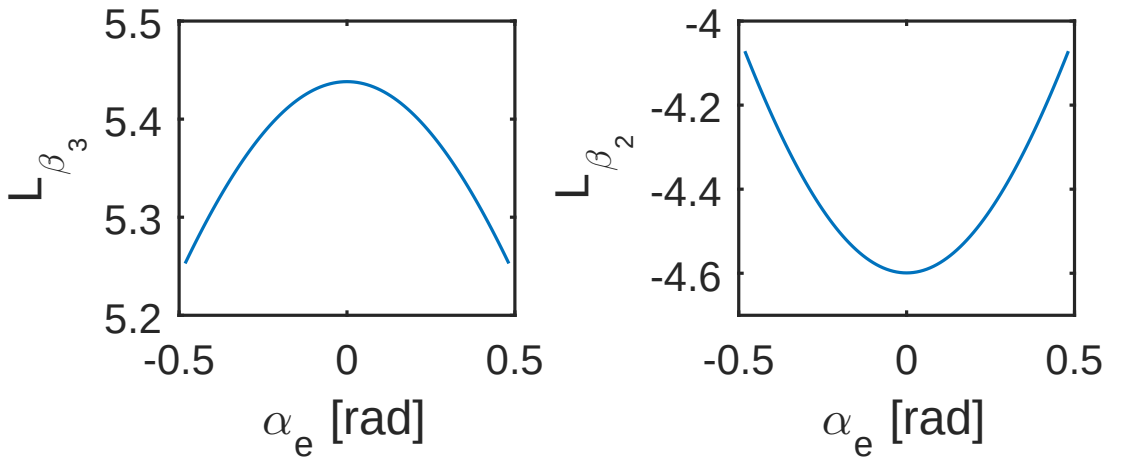

Fig. 4: Optimal feedback gains as a function of linearization point $\alpha_e$. Gains calculated for $Q = 10I$.

The only tunable parameter is the look-ahead distance, $L_r$, where a shorter look-ahead gives a more aggressive tracking but can cause instability while a longer look-ahead gives a smoother tracking but causes bigger tracking offset. By introducing a proportionality term

$$\beta_{3e} = \beta_{3d} + K_p(\beta_{3d} - \beta_3) \tag{15}$$

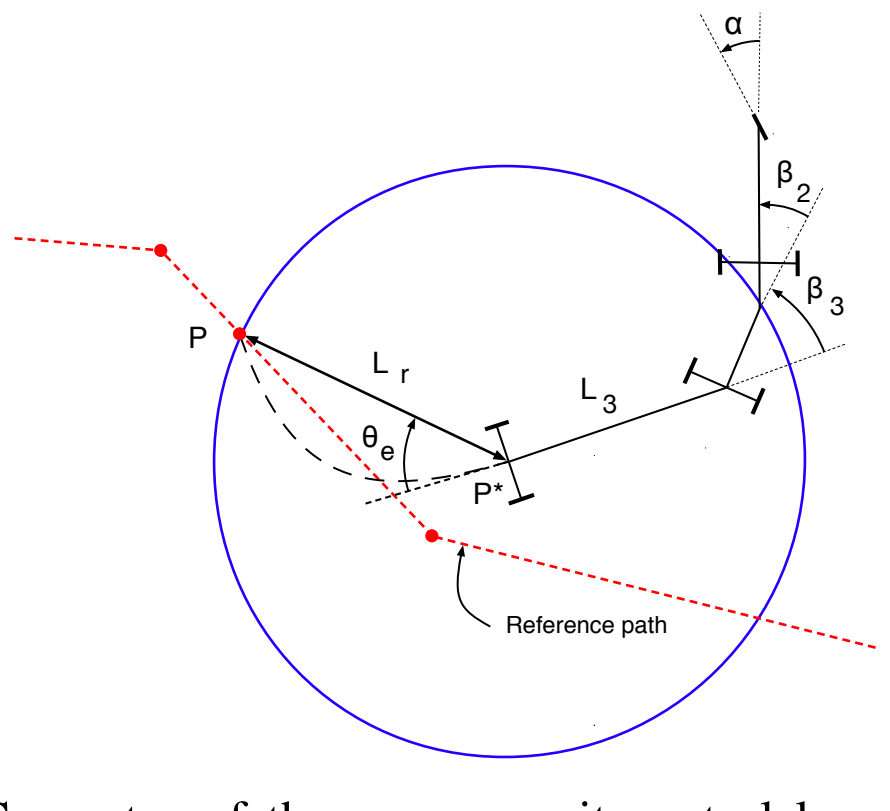

Fig. 5: Geometry of the pure pursuit control law: $L_r$ look-ahead distance, $\theta_e$ angle to look-ahead point and $\alpha$ steer angle.

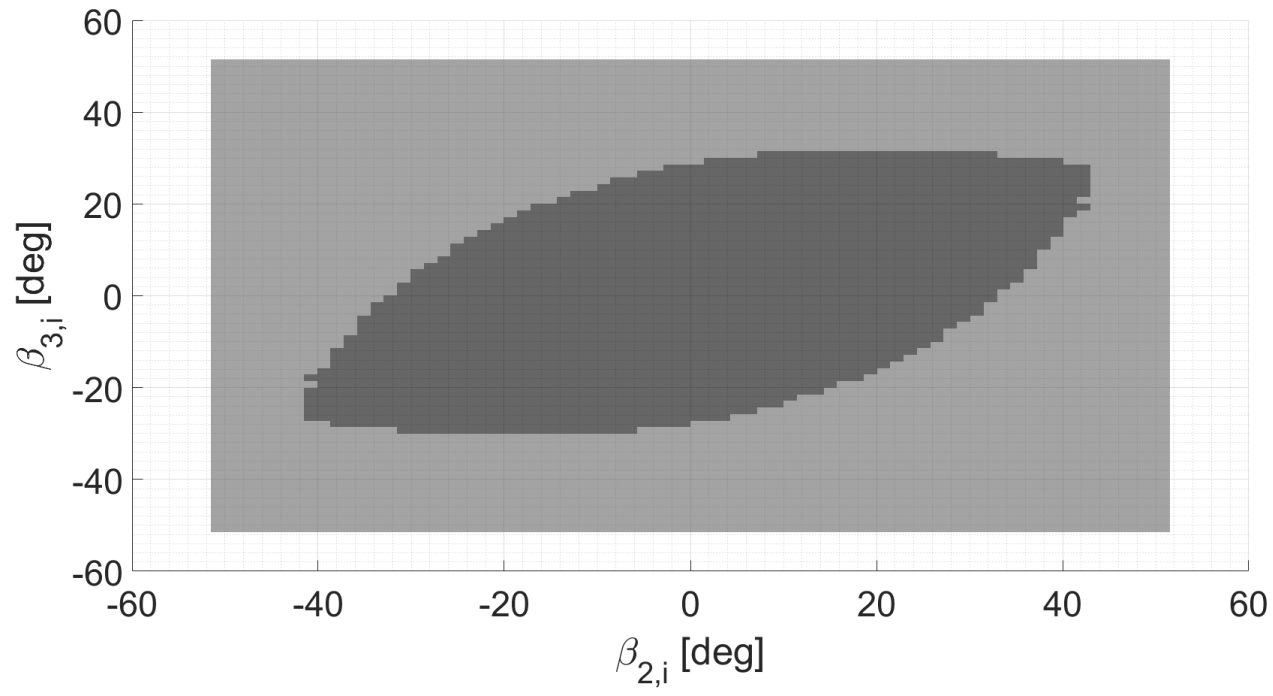

Fig. 6: Simulated region of attraction for different initial configurations of $\beta_2$ and $\beta_3$ for a straight line linearization reference. Light gray region corresponds to unstable initializations and dark gray region corresponds to stable initializations where the truck will converge to the straight line reference.

a more aggressive control can be achieved and experimental results have shown an increase in tracking performance.

### A. Stability

Due to the input constraints it is not possible to globally stabilize the system which might reach a jack-knife state that is impossible to get out of by only reversing. A numerical simulation approach have been used to evaluate convergence from different initial states for $\beta_2$ and $\beta_3$. Given a straight line reference, the parameters for our test platform, $Q = 10I$, $L_r = 50$ cm, $K_p = 0.3$ and $\theta = 0$ the system is simulated from rest and is checked for convergence to produce a region of attraction map as shown in Fig. 6. When the configuration have initial angles with the same sign and is close to an equilibrium point the stability region is quite large but when the angles have opposite sign and is closer to a jack-knife configuration the region shrinks giving the elliptical shape of the region of attraction.

## V. User interface

To make the control system useful for drivers, the piecewise linear reference path that is supposed to be followed by the path tracker need to be created. A path planner could be used to supply the reference but we argue that the available computational resources that can be dedicated to such a task is not enough in the current hardware setup in a modern truck. Instead we present a fast and user friendly approach that could easily be implemented on a touch screen system inside the driver cabin. Assuming we have accurate positioning and some knowledge of the surrounding, e.g. a photographic underlay of the scene or a map created using e.g. ultrasonic sensors, we let the user interactively change the piecewise linear reference path by changing the connection points between the linear elements using a Graphical User Interface (GUI). The system with controller and vehicle model is then simulated along the reference, creating a feasible drivable path that is displayed in real-time on the screen[1]. Since the only requirement on the reference is for it to be piecewise linear, complicated manoeuvres can easily be created using only a few points as seen in Fig. 7. The simulated path can then be sent to the real platform for execution and its feasibility has been ensured by the simulation of the model.

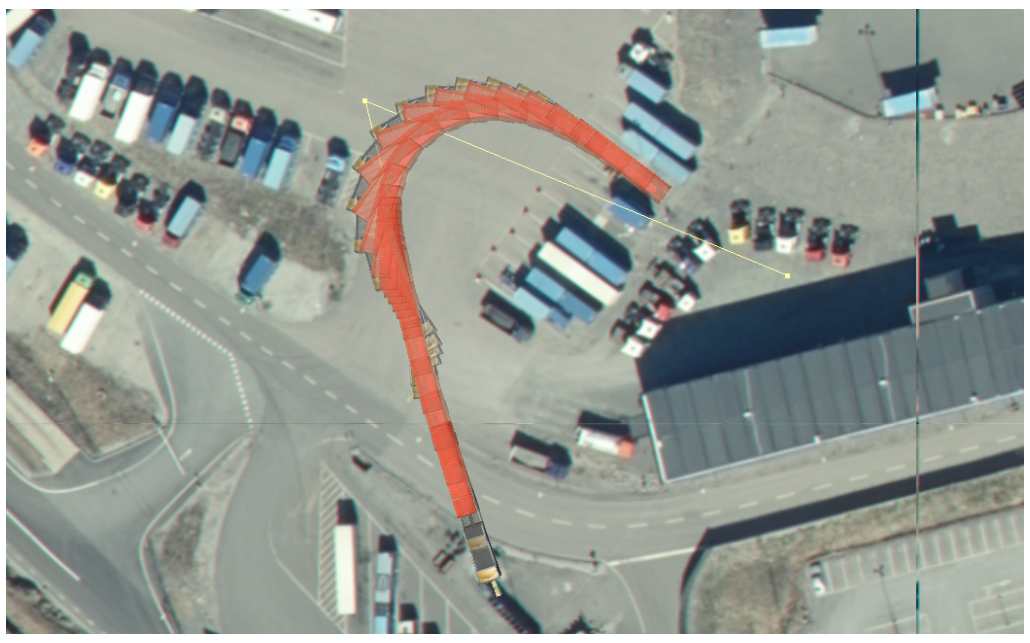

Fig. 7: GUI for creating reference paths for a parking scenario where the driver starts in the lower part of the figure and wants to park the trailer in between the other two trailers next to the end position. The yellow dots connected by the yellow lines represents the reference path. For this manoeuvre the user has only used two control points to create the simulated manoeuvre shown.

## VI. Experimental Platform

The test platform shown in Fig. 1 is used to evaluate the system performance. The platform consist of a small scale truck with Ackerman steering and an off-axle hitching which is connected to a trailer with a turntable dolly. A LEGO NXT control brick is used as the on board computer and the angles $\beta_2$ and $\beta_3$ are measured using two HiTechnic angle sensors. The LQ-controller and the pure pursuit controller is running onboard the NXT control brick with an update frequency of 100 Hz and 10 Hz respectively. A high accuracy Qualisys Oqus motion capture system is used for positioning and the information is relayed to the NXT via a Bluetooth connection. Overall control of the system is done through a laptop computer running the GUI where reference paths can be created and then sent via Bluetooth to the vehicle.

### A. Parameters

By measuring the distances between the wheel axles the following parameters can be found for our vehicle, $L_1 = 19.0$ cm, $L_2 = 14.0$ cm, $L_3 = 34.5$ cm, $M_1 = 3.6$ cm and the steering angle is limited to $\pm 44$ degrees.

## VII. Results

In this section the tracking performance results from both simulation and real experiments are presented. A parking scenario where the GUI is used to plan a path is also studied. The LQ-controller gains used have been found through experiments and the final parameters used for both simulation and real world experiments were $Q = 10I$.

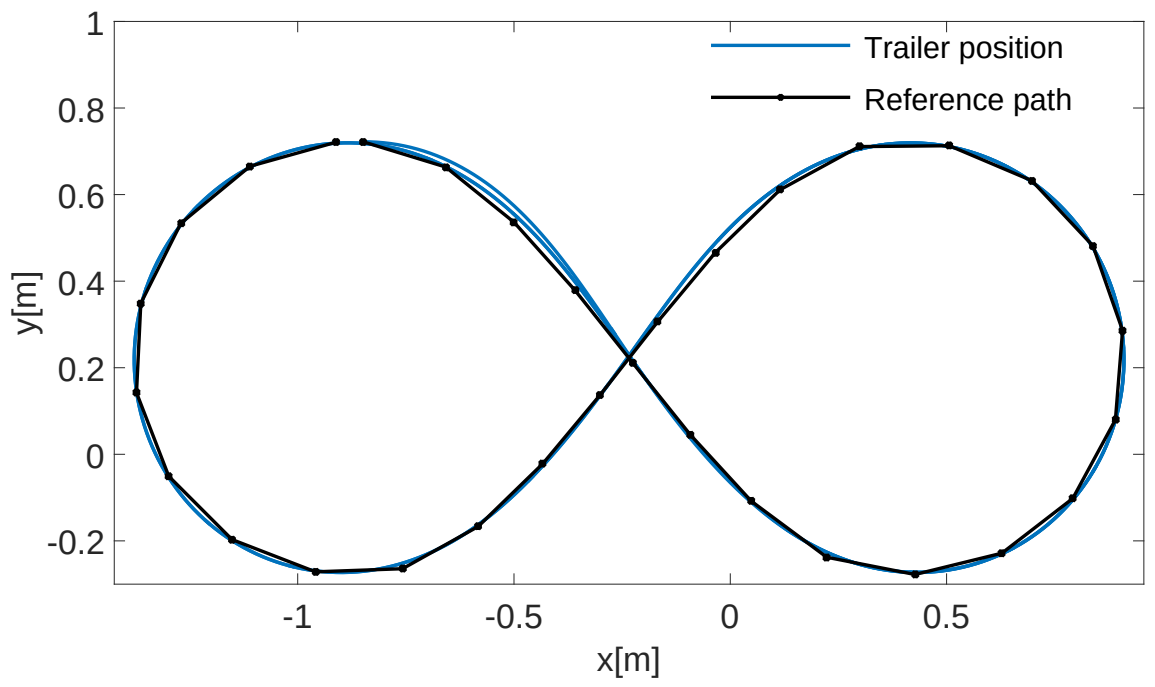

Fig. 8: Simulation result when driving the eight shaped reference path with $L_r = 0.4$ m and $K_p = 0.3$. Blue line is the recorded trailer position and the black dotted line represents the reference path.

### A. Simulation experiments

The tracking performance is first evaluated in a simulation environment where the system can be tested without any unknown disturbances. To challenge the control system a reference in the shape of an eight is constructed to make the vehicle shift between hard left and right turns. The system is then simulated along the reference for five laps using the same specifications as for our test platform and pure pursuit parameters $L_r = 0.4$ m and $K_p = 0.3$. The resulting maximum and mean tracking error in this simulation were 2.81 cm and 0.45 cm, respectively. From Fig. 8 it is seen that the reference is closely tracked and only diverging from the reference a little at the shift between the turns and the straight section. The results are encouraging but measurement errors and the steering backlash is not included in the simulation model, so worse results should be expected for the real world experiments.

### B. Lab experiments

*1) Eight shaped reference:* In the first lab experiment the same reference path as in the simulation experiment is used. In the lab experiments it was found that the pure pursuit parameters that were used for the simulations gave unstable behavior and the look-ahead, $L_r$, had to be increased to 0.5 m to have reliable operation. The vehicle was driven five laps around the course while logging onboard sensors and position from the Qualisys motion capture system. The resulting path is shown in Fig. 9 and the angle measurements for $\alpha$, $\beta_2$ and $\beta_3$ during the run are shown in Fig. 10. From Fig. 9 it is seen that the tracking performance is very consistent and only differ by a few centimeters between the laps and the trailer also tracks the reference quite well but gives a larger error when entering the turns. It can be seen from Fig. 11 that the average tracking error during the five laps is 1.67 cm and the maximum tracking error at any time is 4.15 cm. As expected the tracking error is worse when compared to the simulations and the look-ahead distance even had to be increased to maintain stability, even so a average tracking error of 1.67 cm is acceptable in most situations. The steering mechanism on the test vehicle has a quite significant backlash in the gearing between the steering servo and the wheels which makes the LQ-controller to constantly adjust as can be seen in the plot of the steering angle, $\alpha$, in Fig. 10. A small delay between the $\beta_3$ reference and the $\beta_3$ measurement is seen in Fig. 10 due to the time it takes the LQ-controller to move the system into a new equilibrium configuration. Due to the cascaded nature of the approach this delay can be problematic since it is not explicitly modeled and is one of the drawbacks when path tracking and $\beta_3$ tracking is separated into individual controllers.

[1]Video demonstration of GUI and lab experiments can be found at: `https://youtu.be/4EU-t5_mVmA`

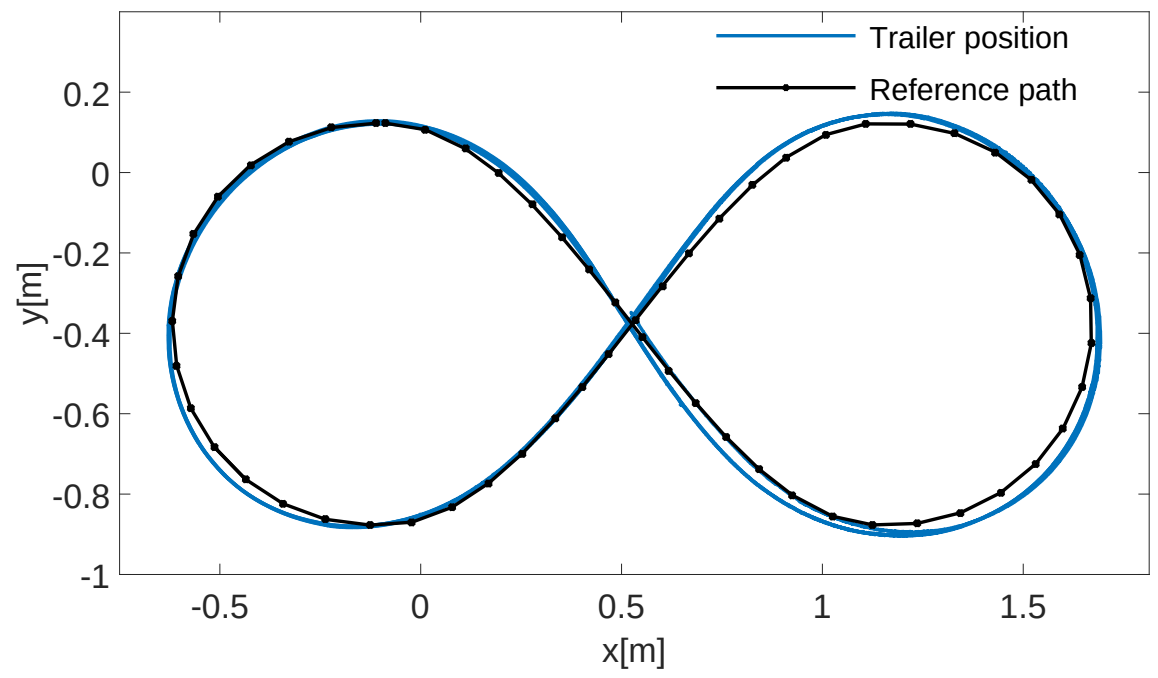

Fig. 9: Tracking results when using the small scale test platform when driving the eight shaped reference path with $L_r = 0.5$ m and $K_p = 0.3$. Blue line is the recorded trailer position and the black dotted line represents the reference path.

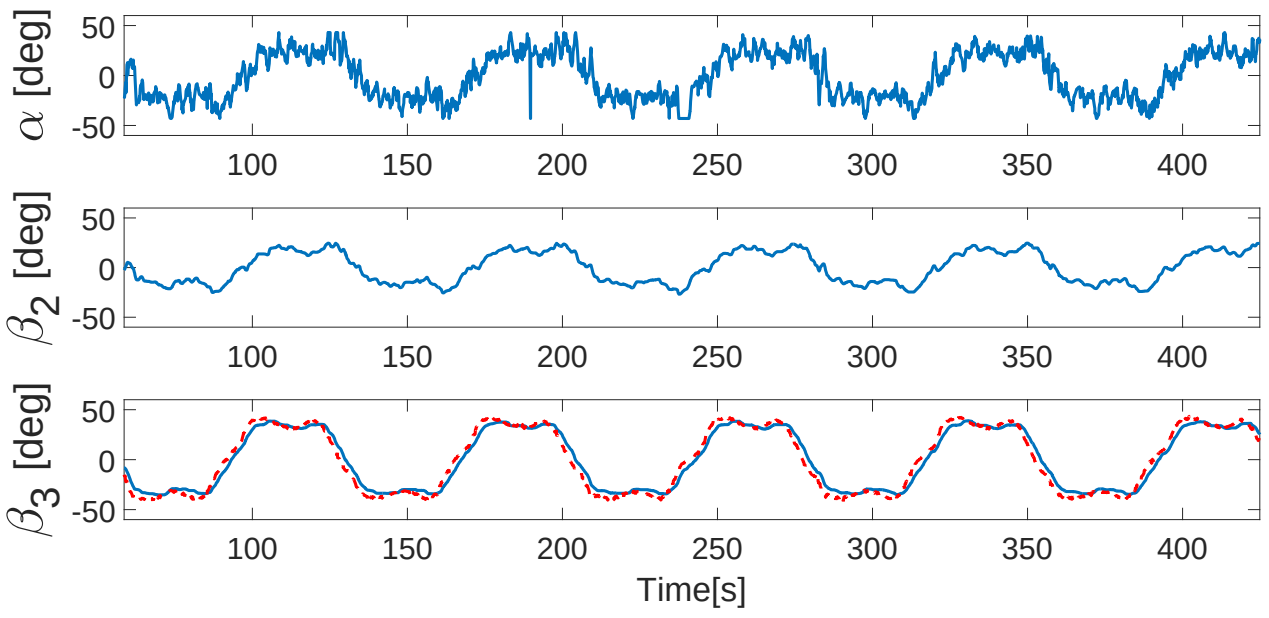

Fig. 10: Top figure: measured steering angle $\alpha$. Middle figure: measured angle $\beta_2$. Lower figure: Blue line shows measured angle $\beta_3$, red dotted line shows $\beta_3$ reference.

*2) GUI reference:* To demonstrate the usefulness of the GUI, the same parking scenario as in Section V is used. In the previous experiments the reference path consisted of only position waypoints for the trailer, but when the simulation model in the reference generation is used the state for the whole configuration is returned. The pure pursuit controller can still only use the simulated path of the trailer as reference but in a parking scenario where we expect the vehicle to behave in the same way as the simulation and other objects should be avoided, it is also important to check the deviation of the truck and dolly against the simulated path. The path shown in Fig. 7 is easily created in the GUI and can then be sent to the platform for execution.

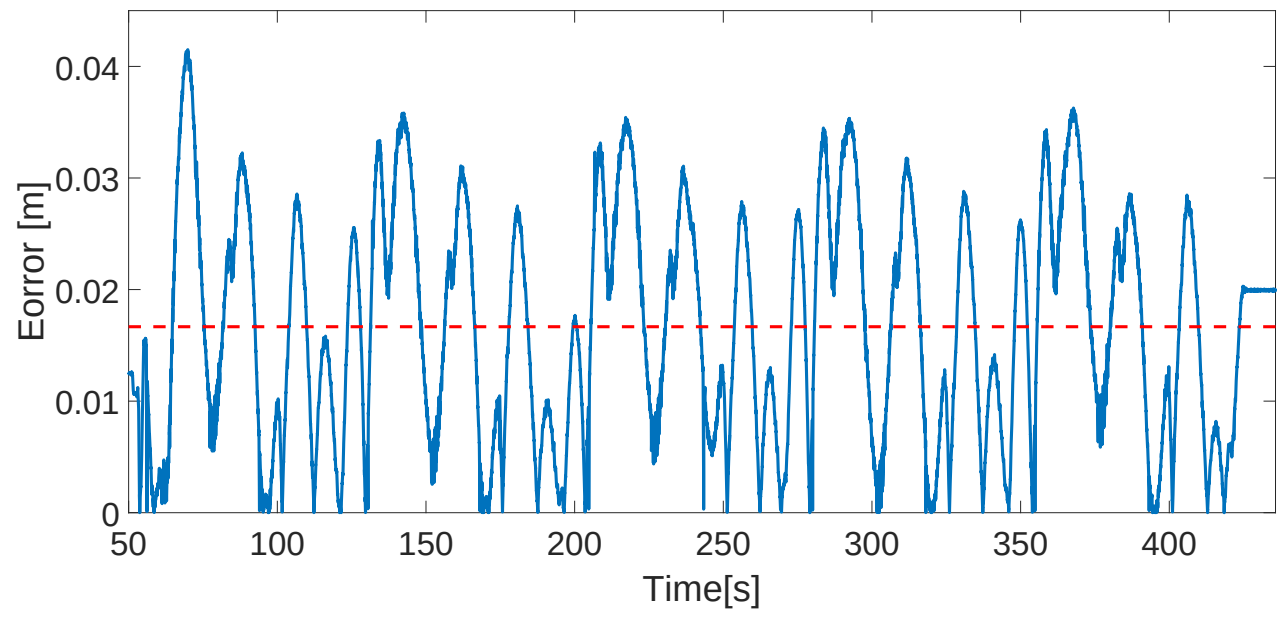

Fig. 11: Blue line represents absolute value of tracking error and red dotted lines shows mean tracking error when driving the eight shaped path.

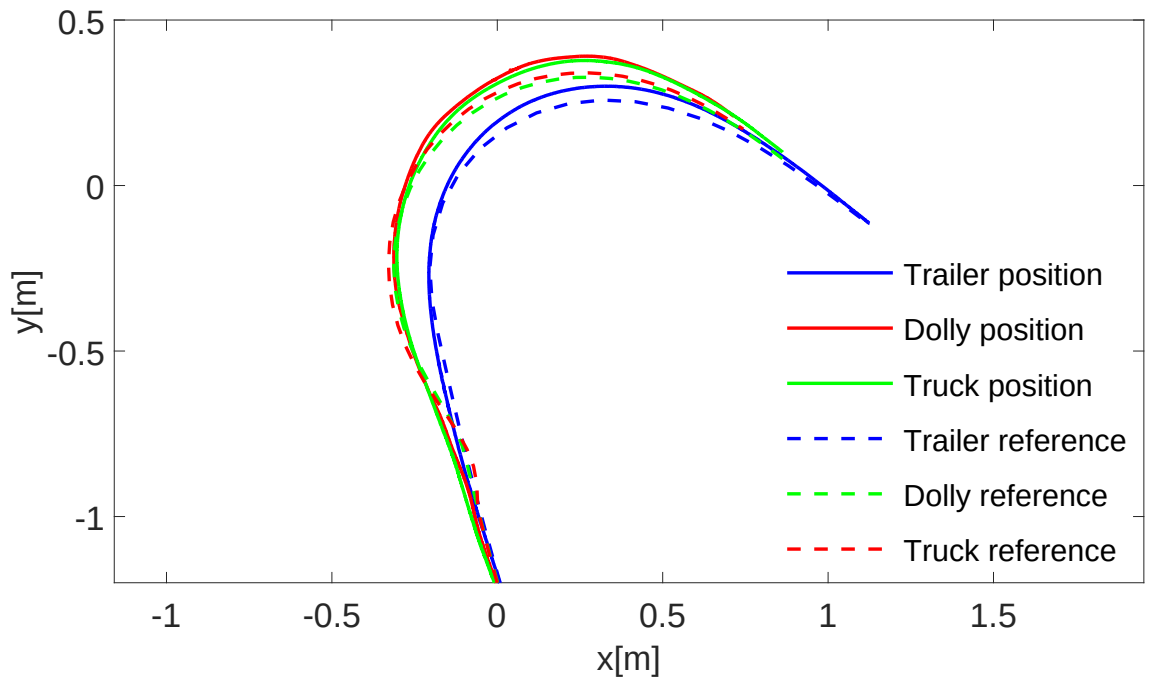

Fig. 12: Path deviation when using the small scale test platform when driving the eight shaped reference path. Blue, red and green lines are the recorded trailer, dolly and truck positions respectively and the blue, red and green dotted lines represents the reference paths.

The reference for trailer, dolly and truck together with the measured positions are shown in Fig. 12. The mean and max deviation from the reference while performing the parking maneuver were $e_{mean} = 2.11$ cm and $e_{max} = 5.10$ cm for the truck, $e_{mean} = 2.13$ cm and $e_{max} = 5.11$ cm for the dolly and $e_{mean} = 1.75$ cm and $e_{max} = 4.25$ cm for the trailer.

## VIII. CONCLUSIONS AND FUTURE WORK

This paper presents a cascaded path tracking and stabilization scheme for a reversing 2-trailer vehicle configuration with off-axle hitching and evaluates the tracking performance in simulation and real world experiments on a small scale test platform. The low level stabilization of the configuration is handled by a gain scheduled LQ-controller that is designed by linearizing the general 2-trailer equations around circular equilibrium configurations, while the path tracking is handled by the well known pure pursuit path tracking algorithm. A maximum and mean tracking error of 2.81 cm and 0.45 cm, respectively, was achieved in the simulations while 4.15 cm and 1.67 cm on the test platform when the system was tested with an eight shaped reference. A GUI was also presented that, given a representation of the surroundings, makes it easy for an operator to manually plan a path for otherwise challenging tasks such as reverse parking manoeuvres. When performing such a manoeuvre it is important to have a small path deviation, not only for the trailer but also for the pushing truck, to avoid collisions with nearby objects. A reversing parking scenario was created in the GUI and then executed by the test platform giving a maximum deviation error of a few centimeters for both the trailer and the truck.

Very general piecewise linear references can be handled by the cascaded approach with a pure pursuit path tracking controller. However, since the controller only concerns the path tracking for the trailer and no explicit tracking is done for the truck and the dolly it could be risky when performing tight manoeuvres. Since the path generation returns the states for the whole configuration, future work will include controller designs that exploit this and minimizes the tracking error for the whole configuration. We will also look into the possibility of using a path planner instead of the manual planning with the GUI and also perform full scale experiments with a full size trailer configuration.

## ACKNOWLEDGMENT

We gratefully acknowledge the Royal Institute of Technology for providing us with the opportunity to perform experiments at their facilities at the Smart Mobility Lab located in Stockholm, Sweden.

## REFERENCES

[1] M. Werling, P. Reinisch, M. Heidingsfeld, and K. Gresser, “Reversing the general one-trailer system: Asymptotic curvature stabilization and path tracking,” *Intelligent Transportation Systems, IEEE Transactions on*, vol. 15, no. 2, pp. 627–636, 2014.

[2] C. Altafini, A. Speranzon, and K. H. Johansson, “Hybrid control of a truck and trailer vehicle,” in *Hybrid Systems: Computation and Control*. Springer, 2002, pp. 21–34.

[3] O. J. Sørdalen, “Conversion of the kinematics of a car with n trailers into a chained form,” in *Robotics and Automation, 1993. Proceedings., 1993 IEEE International Conference on*. IEEE, 1993, pp. 382–387.

[4] D. Tilbury, R. M. Murray, and S. Shankar Sastry, “Trajectory generation for the n-trailer problem using goursat normal form,” *Automatic Control, IEEE Transactions on*, vol. 40, no. 5, pp. 802–819, 1995.

[5] D.-H. Kim and J.-H. Oh, “Experiments of backward tracking control for trailer system,” in *Robotics and Automation, 1999. Proceedings. 1999 IEEE International Conference on*, vol. 1. IEEE, 1999, pp. 19–22.

[6] M. Sampei, T. Tamura, T. Kobayashi, and N. Shibui, “Arbitrary path tracking control of articulated vehicles using nonlinear control theory,” *Control Systems Technology, IEEE Transactions on*, vol. 3, no. 1, pp. 125–131, 1995.

[7] C. Altafini, “The general n-trailer problem: conversion into chained form,” in *Decision and Control, 1998. Proceedings of the 37th IEEE Conference on*, vol. 3. IEEE, 1998, pp. 3129–3130.

[8] R. M. Murray, “Nilpotent bases for a class of nonintegrable distributions with applications to trajectory generation for nonholonomic systems,” *Mathematics of Control, Signals and Systems*, vol. 7, no. 1, pp. 58–75, 1994.

[9] P. Rouchon, M. Fliess, J. Levine, and P. Martin, “Flatness and motion planning: the car with n trailers,” in *Proc. ECC93, Groningen*, 1993, pp. 1518–1522.

[10] P. Bolzern, R. M. DeSantis, A. Locatelli, and D. Masciocchi, “Path-tracking for articulated vehicles with off-axle hitching,” *Control Systems Technology, IEEE Transactions on*, vol. 6, no. 4, pp. 515–523, 1998.

[11] O. Ljungqvist, “Motion planning and stabilization for a reversing truck and trailer system,” 2016.

[12] C. Pradalier and K. Usher, “Robust trajectory tracking for a reversing tractor-trailer system,” 2008.

[13] A. González-Cantos and A. Ollero, “Backing-up maneuvers of autonomous tractor-trailer vehicles using the qualitative theory of nonlinear dynamical systems,” *The International Journal of Robotics Research*, vol. 28, no. 1, pp. 49–65, 2009.

[14] S. Skogestad and I. Postlethwaite, *Multivariable Feedback Control*. Wiley, 2005.

[15] O. Amidi and C. E. Thorpe, "Integrated mobile robot control," in *Fibers' 91, Boston, MA*. International Society for Optics and Photonics, 1991, pp. 504–523.